\documentclass[letterpaper]{article} 
\usepackage{aaai24}  
\usepackage{times}  
\usepackage{helvet}  
\usepackage{courier}  
\usepackage[hyphens]{url}  
\usepackage{graphicx} 

\usepackage{natbib}  
\usepackage{caption} 
\usepackage{algorithm}
\usepackage{algorithmic}
\usepackage{amsmath}
\usepackage{amssymb}
\usepackage{pifont}
\usepackage{soul} 
\usepackage{multirow}
\usepackage{multicol}
\usepackage{subfigure}
\usepackage{threeparttable}

\usepackage{newfloat}
\usepackage{listings}
\DeclareCaptionStyle{ruled}{labelfont=normalfont,labelsep=colon,strut=off} 
\floatstyle{ruled}
\newfloat{listing}{tb}{lst}{}
\floatname{listing}{Listing}

\usepackage[capitalize]{cleveref}
\crefname{section}{Sec.}{Secs.}
\Crefname{section}{Section}{Sections}
\Crefname{table}{Table}{Tables}
\crefname{table}{Tab.}{Tabs.}

\newcommand{\cmark}{\ding{51}}%
\newcommand{\xmark}{\ding{55}}%
\newcommand{\q}[1]{\boldsymbol{#1}} 

\newcommand{\s}[1]{\mathcal{#1}}
\newcommand{\R}{\mathbb{R}}
\newcommand{\F}[1]{\mathrm{#1}}

\newcommand{\onedot}{\@.}
\newcommand{\etal}{{et al}\onedot}
\newcommand{\etalciteyearpar}[1]{\etal~\citeyearpar{#1}}

\title{RadarMOSEVE: A Spatial-Temporal Transformer Network for Radar-Only \\ Moving Object Segmentation and Ego-Velocity Estimation}
\author{
    Changsong Pang\textsuperscript{\rm 1,\rm 2}\equalcontrib,
    Xieyuanli Chen\textsuperscript{\rm 4}\equalcontrib,
    Yimin Liu\textsuperscript{\rm 3},
    Huimin Lu\textsuperscript{\rm 4},
    Yuwei Cheng\textsuperscript{\rm 2,\rm 3}\thanks{Corresponding author}
}

\affiliations{
    \textsuperscript{\rm 1}Northwestern Polytechnical University\hspace{.1in}\\
    \textsuperscript{\rm 2}ORCA-UBOAT\hspace{.1in}\\
    \textsuperscript{\rm 3}Tsinghua University\hspace{.1in}\\
    \textsuperscript{\rm 4}National University of Defense Technology, College of Intelligence Science and Technology\hspace{.1in}\\
    dreamdusty1113@gmail.com,
    chengyw18@tsinghua.org.cn\\
    yiminliu@tsinghua.edu.cn,
    \{xieyuanli.chen,lhmnew\}@nudt.edu.cn

}

\begin{document}

\maketitle

\begin{abstract}
Moving object segmentation (MOS) and Ego velocity estimation (EVE) 
are vital capabilities for mobile systems to achieve full autonomy. Several approaches have attempted to achieve MOSEVE using a LiDAR sensor. However, LiDAR sensors are typically expensive and susceptible to adverse weather conditions. Instead, millimeter-wave radar (MWR) has gained popularity in robotics and autonomous driving for real applications due to its cost-effectiveness and resilience to bad weather. Nonetheless, publicly available MOSEVE datasets and approaches using radar data are limited. Some existing methods adopt point convolutional networks from LiDAR-based approaches, ignoring the specific artifacts and the valuable radial velocity information of radar measurements, leading to suboptimal performance.
In this paper, we propose a novel transformer network that effectively addresses the sparsity and noise issues and leverages the radial velocity measurements of radar points using our devised radar self- and cross-attention mechanisms. Based on that, our method achieves accurate EVE of the robot and performs MOS using only radar data simultaneously. To thoroughly evaluate the MOSEVE performance of our method, we annotated the radar points in the public View-of-Delft (VoD) dataset and additionally constructed a new radar dataset in various environments. The experimental results demonstrate the superiority of our approach over existing state-of-the-art methods. The code is available at https://github.com/ORCA-Uboat/RadarMOSEVE.
\end{abstract}

\section{Introduction}

\begin{figure}
    \centering
    \includegraphics[width=\linewidth]{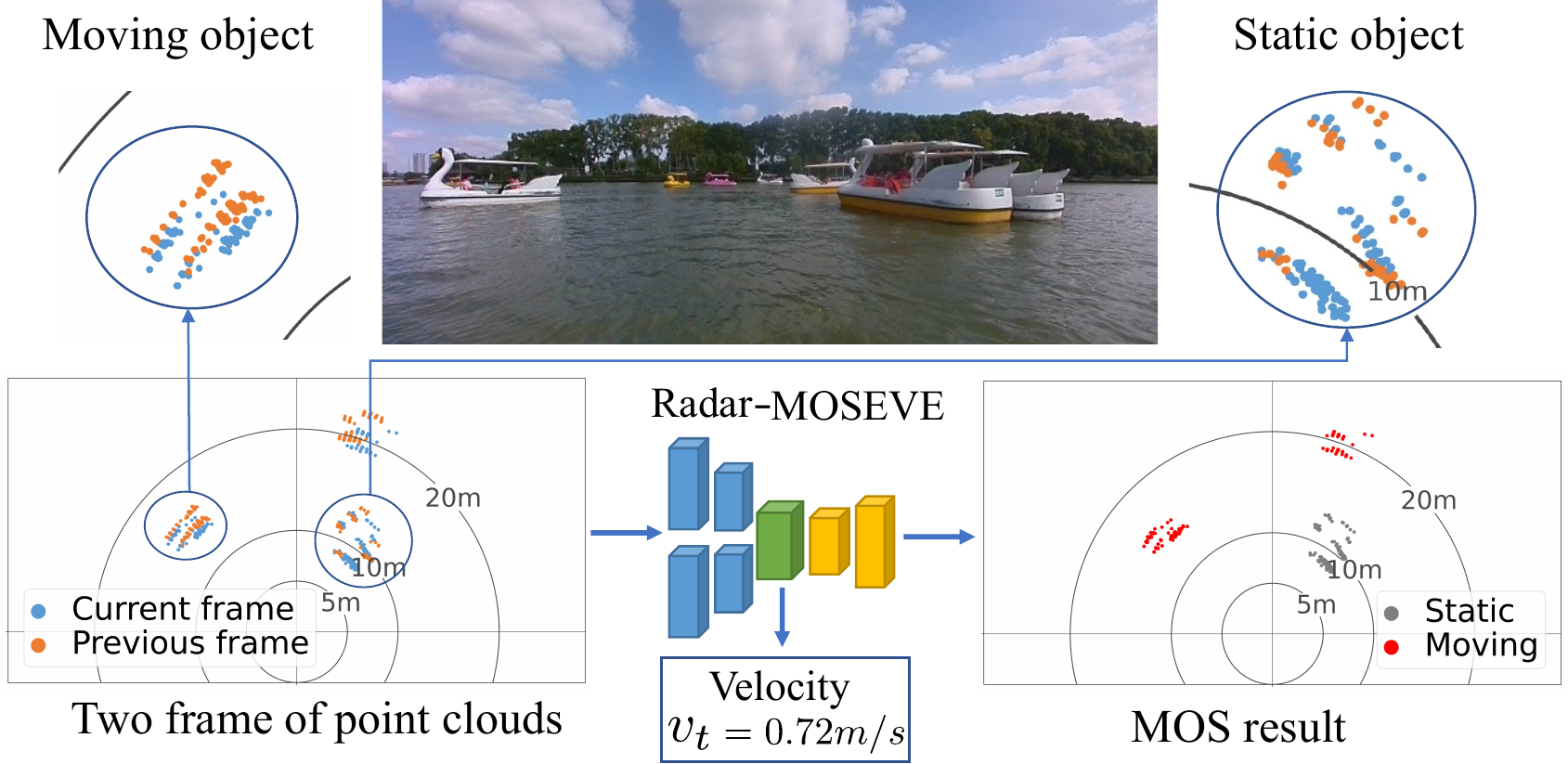}
    \caption{Our MOSEVE network takes two frames of radar point clouds as input and outputs the current ego velocity of the robot. The MOS module takes the velocity-calibrated point clouds to provide the moving segmentation.}
    \label{fig:my_label}
\end{figure}

Simultaneously moving object segmentation (MOS) and ego velocity estimation (EVE) is a challenging task in computer vision, robotics, and autonomous driving. The goal of MOS is to accurately distinguish between moving and static objects in a scene, which is an important component in many downstream tasks, including collision detection, path planning, and navigation. Recently, many studies have been conducted on MOS using LiDAR~\cite{chen2021moving, kim2022rvmos, sun2022efficient, mersch2022receding} and image~\cite{cheng2017segflow, voigtlaender2019feelvos, wang2019learning, patil2021unified} data, which have demonstrated promising performance. However, both LiDAR and camera sensors are susceptible to bad weather conditions, which can cause a substantial reduction in MOS performance when utilized in outdoor real-world applications.   

With the advancement of integrated circuits, 77\,Ghz millimeter-wave radar (MWR)~\cite{sun2020mimo, li2022exploiting} has been applied in mobile robots and autonomous driving~\cite{cheng2022novel}. MWR exhibits strong robustness to different environments and harsh weather conditions~\cite{lu2020see, lu2020milliego}. Additionally, it provides radial velocity information of the measurement point, making it a valuable alternative sensor for MOS tasks. However, few studies have been conducted on MOS using radar data, and directly applying LiDAR MOS methods yields suboptimal results due to the sparsity and noise in the radar points.
Furthermore, existing MOS approaches assume accurate ego-motion estimation from odometry or SLAM systems, which is not always reliable when both the robot and surrounding objects are in motion, making radar MOS tasks highly challenging.

We propose a novel radar point cloud transformer network to simultaneously achieve robust radar MOS and EVE, named RadarMOSEVE. We first modify the self- and cross-attention mechanisms and introduce a novel radar transformer that exploits the velocity information, which is suited for the sparsity characteristics of radar points. The proposed radar transformer is then used to estimate the ego velocity of the robot by utilizing consecutive radar point clouds as input, eliminating the need for odometry estimation from other sources. With the estimated ego velocity, we compensate for the radial velocity of consecutive radar point clouds and input them into a novel MOS module to accurately segment the moving objects in the current observation.

As no public dataset exists for evaluating both MOS and EVE using radar data, we create a novel dataset by utilizing two different platforms in water scenes and ground driving environments. We also annotate the radar point cloud on the public View-of-Delft (VoD)~\cite{palffy2022multi} dataset. We evaluate our method and compare it to existing approaches on both datasets and demonstrate its superiority for radar MOSEVE through comprehensive experiments.

To sum up, our main contributions are threefold:
\begin{itemize}
    \item We propose a novel radar transformer with devised self- and cross-attention mechanisms, which fits the MWR data well and extracts distinctive features from sparse and noisy radar points.
    \item We propose a novel radar MOSEVE framework that fully utilizes the radar Doppler velocity so that the network tackles the MOS and EVE simultaneously.
    \item Our RadarMOSEVE network achieves state-of-the-art performance for two tasks on the VoD dataset and our dataset.
\end{itemize}
\section{Related Work}
\subsection{Moving Object Segmentation}

Many works have been proposed for MOS using image~\cite{voigtlaender2019feelvos, wang2019learning, goel2018unsupervised, patil2021unified} and LiDAR~\cite{chen2021moving, kim2022rvmos, sun2022efficient, gu2022semantics} data. In the vision domain, some methods detect moving objects based on optical flow~\cite{luiten2019premvos, dosovitskiy2015flownet, gong2021flowvos}. For example, Cheng~\cite{cheng2017segflow} fuse target segmentation features with optical flow features to improve MOS accuracy, while Yang~\cite{yang2019unsupervised} propose an adversarial network to check inconsistencies of optical flow to detect moving objects. Other deep network-based methods~\cite{voigtlaender2019feelvos, wang2019learning, goel2018unsupervised, patil2021unified} have also been proposed, such as Goel~\etalciteyearpar{goel2018unsupervised} introduce deep reinforced learning for MOS on image data, and Patilet \etalciteyearpar{patil2020end} propose an end-to-end multi-frame multi-scale encoding-decoding adversarial learning network for segmenting moving objects.

The methods for MOS using LiDAR data can be classified into two main categories: projected range images-based~\cite{chen2021moving, kim2022rvmos, sun2022efficient, gu2022semantics, chen2022automatic} and point cloud-based~\cite{mersch2022receding,he2022empointmovseg,liu2015sparse,mersch2023ral,wang2023iros} methods. The former involves subtracting the range images of consecutive frames to obtain residuals, which are then used to extract spatio-temporal information. Chen~\etalciteyearpar{chen2021moving,chen2022automatic} and Sun~\etalciteyearpar{sun2022efficient} fuse the residual image with range context to enhance the accuracy of LiDAR MOS. The latter category of methods uses sparse convolution to construct a network for segmenting moving point clouds. Some recent works such as Mersch~\etalciteyearpar{mersch2022receding} and He~\etalciteyearpar{he2022empointmovseg} utilize sparse convolution to extract dynamic temporal and spatial features from original point clouds using AR-SI theory~\cite{he2019system}. These features are then employed for MOS using sparse convolution.

Both LiDAR and camera sensors are susceptible to adverse weather conditions, while MWR is a cost-effective and robust alternative. However, there is little research focused on MOS using radar data. One recent study by Zeller~\etalciteyearpar{zeller2022gaussian} utilizes a Gaussian transformer to achieve semantic movable object segmentation in radar data. 
Ding~\etalciteyearpar{ding2022raflow} utilize a self-supervised scene flow for radar MOS and in their later research~\cite{Ding_2023_CVPR}, the performance of MOS is improved by cross modal supervision.

\subsection{Ego-velocity Estimation}
Several ego velocity estimation methods based on 4D MWR have been proposed~\cite{monaco2020radarodo, kellner2013instantaneous, park20213d, steiner2018ego}. Most of them use RANSAC~\cite{kellner2013instantaneous} to estimate the robot's velocity. ICP~\cite{besl1992method} can estimate a transformation between two consecutive point clouds for calculating velocity but cannot work well for noisy radar data. Cen~\etalciteyearpar{cen2018precise} detect landmarks in MWR point clouds to estimate the relative velocity and later~\cite{cen2019radar} perform 3-DOF ego-motion calculation through the separation of key points and graph matching algorithm. Haggag~\etalciteyearpar{haggag2022credible} utilized a probabilistic model without point-to-point correspondence for ego-velocity estimation. However, it relies on static objects for ego-velocity estimation and may not perform well in the presence of moving targets. 
All the above-mentioned methods only focus on ego-velocity estimation and cannot segment the moving objects in radar data.
To the best of our knowledge, our proposed Radar-MOSEVE method is the first work achieving both MOS and EVE using radar data.

\section{Radar Transformer}
\label{sec:radar_cross_transformer}
Before presenting details of our MOSEVE network, we first introduce a novel radar transformer module. With the advent of the transformer network~\cite{vaswani2017attention, devlin2018bert, liu2021swin}, the performance of point cloud segmentation has been significantly improved~\cite{guo2021pct, engel2021point, zhang2022patchformer, zhao2021point}. However, most existing methods focus on LiDAR point cloud processing, and when applied directly to radar point clouds, the performance significantly degrades due to the sparsity of radar data. To overcome this, we propose a novel radar transformer consisting of radar self-attention and cross-attention mechanisms to extract distinctive features on sparse radar data.

\subsection{Radar Self-Attention}
\label{sec:point-transformer}

The original self-attention mechanism in the Point-Transformer (PT)~\cite{zhao2021point} calculates the affinity between a point and its neighbor points. Given one point cloud $\s{P}_t=\{\q{p}_i \in \R^3\}_{i=1}^{N}$, $\q{p}_i=[x_i, y_i, z_i]^{\intercal}$ with $N$ points at time $t$. 
The original PT first uses multilayer perceptrons (MLPs) to generate a D dimensional feature vector $\q{x}_i \in \R^D$ for each point $\q{p}_i$ in $\s{P}_t$. 
K-nearest neighbors (kNN)~\cite{keller1985fuzzy} is then used to find the neighbor point set $\s{Q}_{\q{p}_i}=\{\q{p}_j \in \R^3\}_{j=1}^{K} \subseteq \s{P}_t$ of point $\q{p}_i$, where $K$ denotes the number of neighbor points.
The feature set $\s{X}_{\q{p}_i}$ contains feature vectors of every neighbor point in neighbor point set $\s{Q}_{\q{p}_i}$, and the affinity can be then formulated as:
\begin{equation}
    \q{y}_i=\sum_{\q{x}_j \in \s{X}_i} \F{\theta}(\F{\delta}(\F{\alpha}(\q{x}_i)-\F{\beta}(\q{x}_j)+\omega)) \odot (\F{\gamma}(\q{x}_j)+\omega),
    \label{eq:ptsa}
\end{equation}
where $\F{\alpha}$, $\F{\beta}$ and $\F{\gamma}$ are shared learnable linear transformations, $\F{\delta}$ is an MLP, $\F{\theta}$ is a softmax function and $\omega$ is the position encoding between $\q{p}_i$ and $\q{p}_j$. The position encoding $\omega$ is the residual coordinate between $\q{p}_i$ and $\q{p}_j$ through an MLP with two linear layers and one ReLU nonlinear layer. More details are referred to the original paper~\cite{zhao2021point}.

\begin{figure}[!tb]
\centering
\subfigure[]{
\includegraphics[scale=0.5]{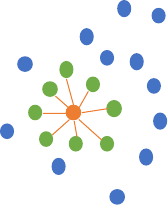}
\label{fig:fig3_2_a}
}
\subfigure[]{
\includegraphics[scale=0.5]{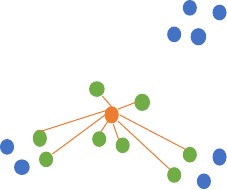}
\label{fig:fig3_2_b}
}
\subfigure[]{
\includegraphics[scale=0.5]{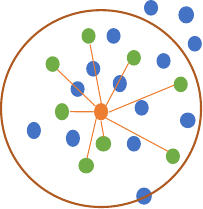}
\label{fig:fig3_2_c}
}
\subfigure[]{
\includegraphics[scale=0.5]{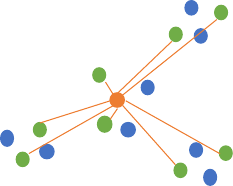}
\label{fig:fig3_2_d}
}
\caption{The orange point is the source point, the green points are the points sampled by the source point and the blue points are the other points. (a) is the sampling result if $k$ is small, (b) is the sampling result if $k$ is large, (c) is the sampling strategy of Object Attention and (d) is the sampling strategy for Scenario Attention.} 
\label{}
\end{figure}

The original point transformer is unsuitable for radar points, since it does not exploit the useful velocity information and the employed kNN can hardly work with sparse and noisy radar point clouds.
As shown in~\cref{fig:fig3_2_a} and~\cref{fig:fig3_2_b}, 
due to the sparsity of radar point clouds, it can be challenging for a single point to find sufficient neighbors belonging to the same object using the original PT. This can lead to misclassifying points from other objects and generating unsuitable features for MOSEVE. To address this issue, we propose a novel radar self-attention mechanism consisting of object attention and scenario attention mechanisms. 
They are designed to better capture the relevant spatial information in radar point clouds and to extract more informative features for MOSEVE.

\begin{figure*}[htb]
    \centering
    \includegraphics[width=0.95\linewidth]{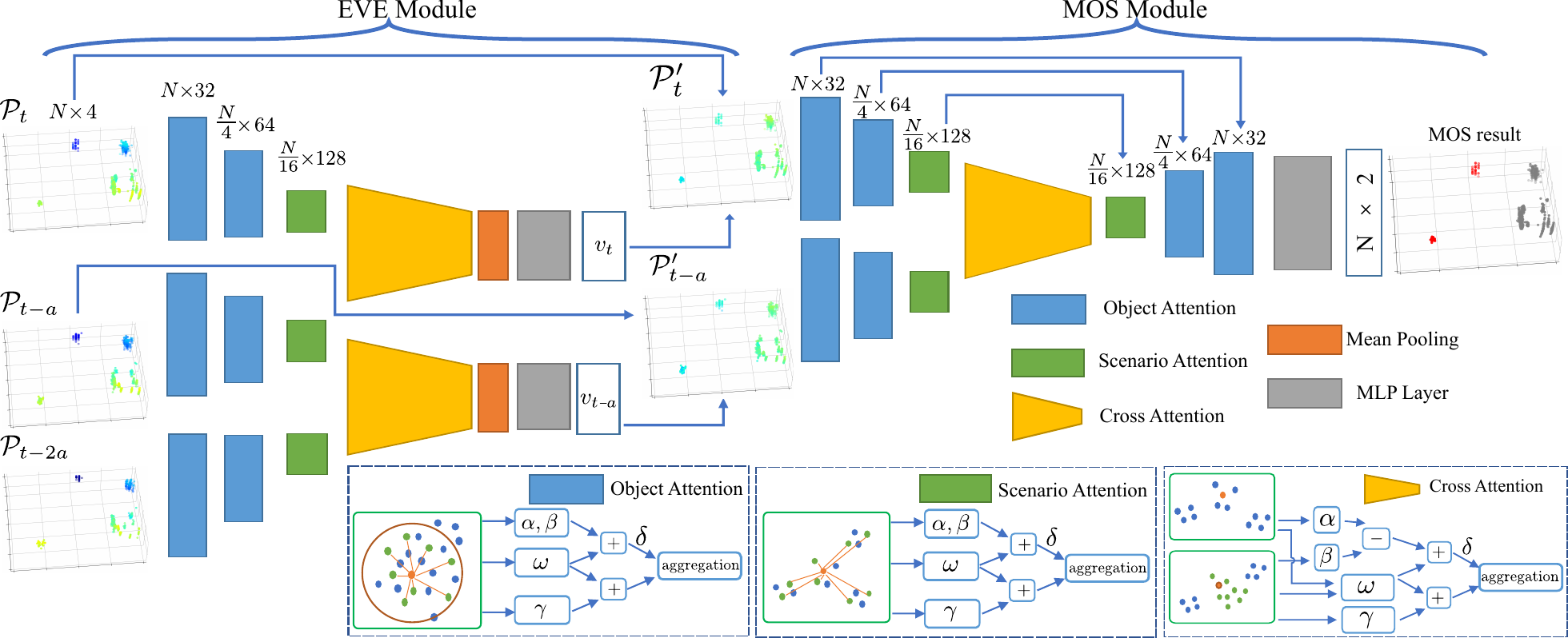}
    \caption{MOS-EVE network for ego-velocity estimation(EVE) and moving object segmentation(MOS module)} 
    \label{fig:fig_3_1}
\end{figure*}

\noindent\textbf{Object Attention.}
We first augment the input with the radial velocity $v_i$ provided by the radar sensor, resulting in a 4D radar point denoted as $\q{p}_i=[x_i, y_i, z_i, v_i]^{\intercal}$. We then introduce an object-size attention operation by exploiting a ball query for the self-attention mechanism to better identify neighbors than simply using K-nearest neighbors in a radar point cloud. As shown in~\cref{fig:fig3_2_c}, a spherical region $\s{Q}_{\q{p}_i}^{\prime}$ with radius $r$ is centered at $\q{p}_i$, and $K$ points are randomly sampled within this region to ensure that each point has the same number of neighbors during network training. If the number of points is less than $K$, some points will be repeatedly sampled. By this, $\q{p}_i$ can attend to all points in the region $\s{Q}_{\q{p}_i}^{\prime}$ after several sampling, providing a relatively stable receptive field that facilitates the understanding of the object that the $\q{p}_i$ belongs to.

\noindent\textbf{Scenario Attention.}
The spatial distribution of static objects remains stable in consecutive frames, while that of moving objects can vary largely. Consequently, the relationships between points in the scenario can help reason about motion. However, the original self-attention in PT and object attention have limited receptive fields, thus unable to capture inter-object information. Therefore, we propose scenario-level self-attention for a radar point cloud $\s{P}_t$. We perform interval sampling on $\s{P}_t$ based on the distance to $\q{p}_i$ to create the point set $\s{Q}_{\q{p}_i}^{\prime\prime}$, which expands the receptive field of $\q{p}_i$ as depicted in~\cref{fig:fig3_2_d}. This process allows the embedding of $\q{p}_i$ to incorporate features from many other objects. After multiple downsampling $\s{P}_t$ using the farthest point sampling (FPS)~\cite{moenning2003fast}, the scenario attention generates a feature that encapsulates the relationships between the object of $\q{p}_i$ and other objects in the scene.

By leveraging both object and scenario attention, our proposed radar self-attention mechanism effectively extracts valuable features from sparse and noisy radar points. 
The former enhances feature sharing within individual object points, improving object motion feature extraction from sparse radar points. The latter captures scene-level features for each point by analyzing the scene's point characteristics.

\subsection{Radar Cross-Attention}
\label{sec:radar_cross}
To fully exploit the spatio-temporal information of 4D radar point clouds, it is natural to utilize sequential data from consecutive frames for ego velocity and MOS estimation. Unlike existing work using sparse convolutions~\cite{mersch2022receding}, in this work, we propose a novel radar cross-attention mechanism to effectively capture the spatio-temporal dependencies among two radar point clouds.
We take the previous frame $\s{P}_{t-a}=\{\q{p}_k \in \R^4 \}_{k=1}^{M}$ together with the current frame $\s{P}_t$ as the input of the proposed radar cross-attention, where $a$ denotes the time interval between $\s{P}_{t-a}$ and $\s{P}_{t}$. 
We use the proposed ball sampling to sample $K$ points from $\s{P}_{t-a}$ to form the neighbor point set $\s{S}_{\q{p}_i} =\{\q{p}_j\}_{j=1}^{K} \subseteq \s{P}_{t-a}$ of $\q{p}_i \in \s{P}_{t}$. Then, $\q{p}_i$ and $\q{p}_j$ are transformed into the embedding features $\q{y}_i$ and $\q{y}_j$ by our radar self-attention module. 
$\s{Y}_{\q{p}_i}=\{\q{y}_j\}_{j=1}^{K}$ is the feature embedding set of all points in $\s{S}_{\q{p}_i}$.
We conduct  cross-attention on $\q{y}_i$ and $\q{y}_j$ as:
\begin{equation}
    \q{z}_i=\sum_{\q{y}_j \in \s{Y}(i)} \F{\theta}(\F{\delta}(\F{\alpha}(\q{y}_i)-\F{\beta}(\q{y}_j)+\epsilon)) \odot (\F{\gamma}(\q{y}_j)+\epsilon),
    \label{eq:ca}
\end{equation}
where $\alpha$, $\beta$, $\gamma$, $\delta$, and $\theta$ are as same linear function as those in~\cref{eq:ptsa}, but with different parameters. $\epsilon$ is the position encoding between $\q{p}_i$ and $\q{p}_j$, which is the residual coordinate between $\q{p}_i$ and $\q{p}_j$ through an MLP. 
The MLP structure of EVE differs from MOS. EVE has two linear layers and one ReLU, while MOS has three linear layers and three ReLU nonlinear layers since MOS is more challenging and needs large models to learn.

\section{Radar MOSEVE Network}
\subsection{Overview}
We aim to achieve reliable and accurate MOSEVE simultaneously using MWR point clouds. To this end, we propose a novel multi-level attention-based network based on our proposed radar transformer modules, as depicted in~\cref{fig:fig_3_1}. The network comprises two main modules, the EVE module and the MOS module, designed to accomplish both tasks. The EVE module employs two frames of 4D radar point clouds $\s{P}_t$ and $\s{P}_{t-a}$ to estimate the robot's ego velocity $v$, as detailed in~\cref{sec:EVE}. To improve the accuracy of the ego-velocity estimation, we propose a novel Doppler loss during training. In the MOS module outlined in~\cref{sec:mos}, we first compensate the radial velocity of radar point clouds using the EVE results, ensuring that the radial velocity of points on surrounding static objects is close to zero. We refer to the velocity-calibrated point clouds as $\s{P}_{t}^{\prime}$ and their previous frame as $\s{P}_{t-a}^{\prime}$. Our MOS module takes these calibrated point clouds as input and segments the moving objects in the current frame. The training process is detailed in~\cref{sec:pipeline}.

\subsection{Ego-velocity Estimation}
\label{sec:EVE}

\textbf{EVE Backbone.} 
It is composed of four stages. The first three stages extract intra-frame features from two radar point clouds, $\s{P}_{t-a}$ and $\s{P}_{t}$ using our radar self-attention module. It first applies the object attention module at two different resolutions to extract object features for each point in two point clouds. Then, the scenario attention module aggregates the features of different objects in each radar observation. At the final stage, the radar cross-attention module is applied to fuse point features in two point clouds and generate the inter-frame feature by incorporating spatio-temporal information from the two radar point clouds. As the network deepens, the point clouds are gradually downsampled with sampling rates of [1, 4, 4, 1]. Thus, the point set for each stage is [$\text{N}_\text{p}, \text{N}_\text{p}/4, \text{N}_\text{p}/16, \text{N}_\text{p}/16$], where $\text{N}_\text{p}$ is the number of input points. We utilize the FPS method to obtain a well-spread downsampled subset of the point cloud.

\noindent\textbf{EVE Head.} 
In the EVE head, we apply global average pooling to obtain a 128-dimensional global feature vector. This vector is then passed through an MLP consisting of three linear layers and two ReLU non-linear layers to predict the robot's velocity. The output sizes of the MLP layers are [256, 64, 1]. The final one-dimensional output represents the robot's velocity in that frame.

The velocity of static objects in the environment should be zero. However, as the robot moves, the raw velocity measurements of radar points are relative to the robot. 
Suppose the robot moves forward at a velocity of $v$. We project it in the direction of $\q{p}_i$, obtaining the radial velocity as:
\begin{equation}
    \centering
    \begin{aligned}
        \hat{v}_i = -v \cdot \frac{y_i}{\sqrt{x_i^2+y_i^2+z_i^2}}.
    \end{aligned}
\end{equation}

When $\q{p}_i$ belongs to a static object, its radial velocity $v_i$ measured by radar should be equal in magnitude to the projection velocity $\hat{v}_i$ of the robot's ego velocity $v$.
Therefore, enforcing the radial projection of the EVE network output to be equal in magnitude to the radial velocity of the static point allows the network to learn the ego velocity of the robot. We present more training details in~\cref{sec:pipeline}.

\subsection{Moving Object Segmentation}
\label{sec:mos}
\textbf{Velocity Compensation.}
Assuming the robot moves forward in a short time interval $a$, it is necessary to convert the velocity measurement of a 4D radar point to the global coordinate system before conducting radar MOS. Because the raw velocity measurement is relative to the robot motion, it may confuse the network in determining the motion state of a point. To tackle this, we use our EVE network output $\hat{v}$ to compensate for the radial velocity $v_i$ of the point cloud, bringing the absolute velocity of static points close to zero.
The calibrated velocity $v_i^{\prime}$ of $\q{p}_i$ is calculated as
\begin{equation}
    \centering
    \begin{aligned}
        v_i^{\prime} = \hat{v} \cdot \frac{y_i}{\sqrt{x_i^2+y_i^2}} - v_i \cdot \frac{\sqrt{x_i^2+y_i^2+z_i^2}}{\sqrt{x_i^2+y_i^2}}.
    \end{aligned}
    \label{eq7}
\end{equation}
Since pitch, roll, and z-axis changes are typically small for ground and water vehicles in a short time, we calculate the radial velocity of the point cloud in the XOY plane.

After velocity calibration, the radial velocity of static points is approximately zero, while some moving points may also have velocities close to zero due to measurement noise and movement direction. Therefore, determining the motion state of a radar point cloud based solely on radial velocities is challenging. Thus, we use the calibrated point clouds 
$P^{\prime}_t=\{[x_i, y_i, z_i, v_i^{\prime}]\}_{i=1}^N$ 
and $P^{\prime}_{t-a}$ as input data for our MOS network to reason about the motion of objects.

\noindent\textbf{MOS Backbone.}
We use a U-net design for the MOS network using the same encoder structure as that of our EVE module. For decoding, we use trilinear interpolation to upsample the features to high-resolution point clouds. These features are concatenated with the corresponding resolution encoding features via a skip connection. We use object attention and scenario attention in different resolutions to fuse the different features, while different modules are connected during the decoding stage via the transition-up module~\cite{zhao2021point}. The output of the module is a moving segmentation feature of size (N,32).

\noindent\textbf{MOS Head}. 
Given the moving segmentation feature for each point in the point cloud $\s{P}_t^{\prime}$, we use an MLP consisting of three linear layers and two ReLU nonlinear layers to convert these features into the final logits for MOS.

\subsection{Network Training}
\label{sec:pipeline}
We first train our EVE module, then use the EVE estimates to compensate for the radar point velocities, and finally train the MOS module.
The EVE training loss $L_{EVE}$ consists of an MSE loss $L_{mse}$ and proposed new Doppler loss $L_{dop}$ as
\begin{equation}
\begin{aligned}
    L_{EVE} = \  L_{dop} + L_{mse}.
\end{aligned}
\end{equation}

Given the points $\q{p}_i$ of all static objects form the point set $\s{P}_s = \{\q{p}_i \in \R^4 \}_{i=1}^{N_s} \subseteq \s{P}_t$ using our MOS labels, the Doppler loss is defined as
\begin{equation}
    \centering
        L_{dop} = \frac{1}{N_s} \sum_{i=1}^{N_s} \left| \hat{v} \cdot \frac{y_i}{\sqrt{x_i^2+y_i^2+z_i^2}} - v_i \right|,
    \label{eq:dop}
\end{equation}
which makes the output $\hat{v}$ of EVE network equal to the robot ego velocity $v$. 
MSE loss $L_{mse} = \ \frac{1}{N_b}\sum^{N_b} (v - \hat{v})^2$ is used to provide additional supervision to the EVE estimates by comparing them to the ground truth velocity. 

The calibrated 4D radar point clouds and the corresponding MOS labels are then used to train the MOS module with a typical weighted cross-entropy loss 
\begin{equation}
L_{mos} = -~\sum_{c=1}^{C}{w_c\,l_c\,\log{\big(\hat{l}_c\big)}} ,
\end{equation}
where $C$ contains moving and static two classes, $w_c$ is the weight factor for class $c$, $\hat{l}_c$ and $l_c$ are the predicted and ground truth labels.

\section{Experimental Results}
In this section, We present our experiments to illustrate that our approach is able to achieve both MOS and EVE using only radar data and outperforms existing state-of-the-art methods on our dataset and the VoD dataset. In addition, we provide multiple ablation studies to experimentally validate the effectiveness of all our network designs and architecture.

\subsection{Experimental Setup}
\textbf{Dataset.} We evaluate our methods on our dataset, including  13,654 frames of point clouds. We perform moving object segmentation using LiDAR data and then label the corresponding radar points as moving. To ensure accuracy, we manually verify and correct the labels. In addition, we also validate the advancement of our method on the open-sourced VoD dataset. The MOS label created by \cite{Ding_2023_CVPR} exists lots of errors, so we annotate the dataset again. We choose the suitable data and re-split them to trainval and test datasets. The final dataset contains about 3,000 frames. More details of datasets are provided in the supplementary.

\noindent\textbf{Metrics.} For MOS, intersection-of-union(IoU) is calculated as a MOS metric to compare with other baselines. Since MOS is a binary classification task for each point, We also use F1-Score and accuracy to comprehensively evaluate the MOS performance. For EVE, the mean absolute error(MAE) and mean square error(MSE) of velocity metric the accuracy and robustness of EVE well. On some scenarios, high precision is more effective for path planning and robotic odometry. Therefore, we compare all methods with several precision results at different ego velocity thresholds, i.e. the ratio of estimates with errors smaller than a threshold.

\noindent\textbf{Implementations Details.}
In all experiments, we set the time interval $a=10$, the number of neighbors $k=16$ in our radar transformer, and the sampling interval $g=2$ in scenario attention. We randomly sample 512 points of every frame to test the model performance.
We implement our Radar-MOSEVE network in PyTorch. We use the Adam optimizer with a weight decay of 0.001 to train our network of two tasks. The batch size is set to 4. For EVE, we train for 60 epochs. For MOS, we train for 50 epochs. The initial learning rate is 0.001. The learning rate decay ratio is set at 0.5, and it occurs every 10 epochs for MOS and 20 epochs for EVE. We train our model on a GTX3060 GPU, taking 12 hours in total. For all experiments, we use the same seed and take the average results of our trained models three times.


\begin{figure*}[t]
\centering
\includegraphics[width=0.94\linewidth]{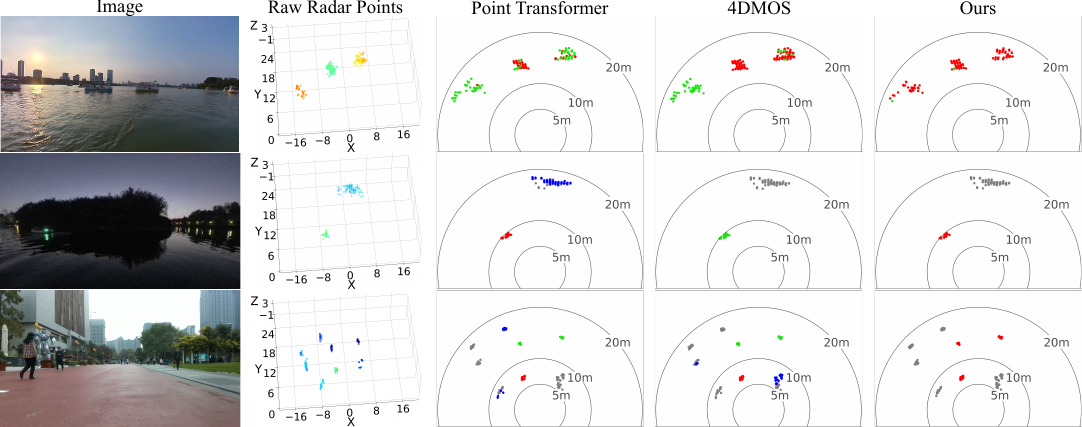}

\caption{Qualitative results of PT, 4DMOS and ours. Red points are the true moving points, gray points are the true static points, green points are the false moving points, and blue points are the false static points. Better view with colors.}
\label{fig:result}
\end{figure*}

\begin{table*}[!tb]	
\centering
\addtolength{\tabcolsep}{0.5pt}
\renewcommand\arraystretch{1.1}
\setlength{\tabcolsep}{9pt}
\small
\begin{tabular}{l|ccc|ccc|ccc}
\hline
\multirow{2}{*}{\textbf{Method}} &
\multicolumn{3}{c|}{\textbf{IoU}$\uparrow$}&
\multicolumn{3}{c|}{\textbf{F1}$\uparrow$} &
\multicolumn{3}{c}{\textbf{Acc}$\uparrow$} \\
 & {\textbf{Static}}& {\textbf{Moving}} & {\textbf{Avg}}   & {\textbf{Static}} & {\textbf{Moving}} & {\textbf{Avg}}  & {\textbf{Static}} & {\textbf{Moving}} & {\textbf{Avg}} \\

\hline 
ICP~\citeyearpar{besl1992method}  & 23.6& 26.7& 25.2  & 33.8 & 38.3 & 36.1 & 48.9 & 39.3 & 44.1  \\
RANSAC~\citeyearpar{kellner2013instantaneous} & 36.2 & 29.1 & 32.6   & 49.0 & 40.7 & 44.9 & 55.4 & 44.8 & 50.1   \\
PT~\citeyearpar{zhao2021point}  & 62.7 & 46.9 & 54.8  & 72.5 & 54.2 & 63.4 & 74.2 & 70.3 & 72.2  \\
Stratified-Transformer~\citeyearpar{lai2022stratified} & 58.9 & 54.5 & 56.7   & 74.1 & \underline{70.6} & 72.3   & 74.3 & 71.0 & 72.6   \\
Point-BERT~\citeyearpar{yu2022point} & 59.9 & 45.1 & 52.5   & 70.1 & 51.5 & 60.8   & 77.5 & 62.8 & 70.1   \\
4DMOS~\citeyearpar{mersch2022receding} & 67.1 & 54.6 & 60.8   & 76.5 & 62.3 & 69.4   & 78.3 & 73.2 & 75.8   \\
\hline
PT+V  & 66.3 & \underline{56.7} & 61.5  & 74.1 & 63.9 & 69.0 & \underline{81.7} & \underline{72.5} & 76.2  \\
4DMOS+V & \underline{67.9} & 55.4 & \underline{61.7}   & \underline{76.6} & 62.5 & \underline{69.6}   & \textbf{85.0} & 66.7 & \underline{76.0}   \\
\hline 
Ours & \textbf{73.3} & \textbf{67.2} & \textbf{70.2} & \textbf{79.8} & \textbf{73.2}  & \textbf{76.5} &81.4 & \textbf{82.5} & \textbf{81.9}  \\
\hline
\end{tabular}
\begin{tablenotes}
\footnotesize
    \item Bold numbers indicate the best performance while the underlined ones are the second best. +V represents using additional radial velocity.
\end{tablenotes}
\caption{The MOS experimental results on our dataset.}
\label{tab:4_3_mos}
\vspace{-0.2cm}
\end{table*}

\subsection{Evaluation on MOS}
In the MOS task, we compare our method against both traditional and deep learning-based approaches using the IoU, F1 and Accuracy. These metrics are calculated separately for the static and moving classes, as well as for the average scores. For traditional methods, we utilize the point-to-point ICP algorithm~\cite{besl1992method} to compute the transformation between two radar point clouds. We then use position residuals with a threshold to identify moving and static points. Once we obtain the EVE of the robot, we can check the difference between the EVE results and radial velocities of the point raw measurement against a threshold to determine moving and static points. Therefore, we use RANSAC~\cite{kellner2013instantaneous} as another baseline by employing it to calculate the radial velocity of the robot.
As for deep learning methods, 
We adopt Point Transformer (PT)~\cite{zhao2021point} as our baseline, as it has demonstrated good performance in other point cloud semantic segmentation tasks. We also compare our method with other high-performing transformer-based methods, Stratified-Transformer~\cite{lai2022stratified} and Point-BERT~\cite{yu2022point}. Additionally, we adopt the state-of-the-art LiDAR method 4DMOS~\cite{mersch2022receding} as our baseline, and also evaluate the performance of PT and 4DMOS with additional radial velocity information.

As shown in~\cref{tab:4_3_mos}, our method significantly outperforms all baseline methods across all evaluated metrics on average over all classes. Particularly in terms of mIoU, our method achieves 70.2\% and surpasses the state-of-the-art 4DMOS by 9.4\%. Additional radial velocity information can improve the performance, while our method still outperforms the improved baselines, especially in moving objects with more than 10\,\% improvement. \cref{fig:result} shows the qualitative results on the test set of different methods. Compared to other methods, our approach demonstrates superior capabilities in accurately segmenting the edges of objects and multiple objects in the entire scene into moving and static. 

\begin{table}[!tb]	
\centering
\addtolength{\tabcolsep}{0.5pt}
\renewcommand\arraystretch{1.1}
\setlength{\tabcolsep}{2.4pt}
\small
\begin{tabular}{l|c|c|ccc}
\hline
\multirow{2}{*}{\textbf{Method}} &
\multirow{2}{*}{\textbf{MAE}$\downarrow$} &
\multirow{2}{*}{\textbf{MSE}$\downarrow$} &
\multicolumn{3}{c}{\textbf{Precision$\uparrow$}} \\
& & & {\textbf{$<$0.1m/s}}& {\textbf{$<$0.3m/s}}& 
{\textbf{$<$0.5m/s}}  \\

\hline 
ICP~\citeyearpar{besl1992method}  & 0.842 &  0.870 & 1.1 & 6.9 & 25.2 \\
RANSAC~\citeyearpar{kellner2013instantaneous}  & 0.601 & 0.531 & 10.2 & 25.9 & 49.6 \\
PT~\citeyearpar{zhao2021point}  & \underline{0.330} & \underline{0.175} & \underline{21.5} & \underline{52.8} & \underline{76.5} \\
\hline 
Ours
& \textbf{0.182} & \textbf{0.065} &\textbf{43.3} &\textbf{79.7} &\textbf{94.3}\\
\hline
\end{tabular}
\caption{The EVE experimental results on our dataset. }
\label{tab:4_4_ve}
\normalsize
\vspace{-0.2cm}
\end{table}

\subsection{Evaluation on EVE}
For the EVE tasks, we compare our method against both traditional and deep learning-based approaches using MAE and MSE of velocity estimates. We additionally provide the precision results at different ego velocity thresholds.
As mentioned, we take RANSAC~\cite{kellner2013instantaneous} as a baseline in estimating the radial velocity measured by radar points. The transformation generated by ICP~\cite{besl1992method} can be used to calculate the velocity.  
We also train a PT~\cite{zhao2021point} for EVE as a learning-based baseline. 

The EVE results are presented in~\cref{tab:4_4_ve}. As shown, our method attains the smallest MAE and MSE in EVE, which outperforms all the baseline methods. Especially for precision with a threshold of 0.5\,m/s, our method achieves 94.3\% surpasses PT by more than 17\%. Our method consistently performs stable EVE even when the vehicle moves fast.

\subsection{Evaluation on VoD dataset}
To verify the generalization of our method, we evaluate our method on the VoD dataset. Apart from the methods mentioned earlier, we also compare our methods with radar-based deeping learning methods, RaFlow~\cite{ding2022raflow} and CMFlow~\cite{Ding_2023_CVPR}. We only compare them on the VoD dataset, because the label of scene flow is necessary for the training stage of their networks. 

Due to page limitation, we provide more experiments and qualitative results on VoD dataset in the supplementary. The experimental results indicate our method obtains competitive results both on MOS and EVE tasks in the VoD dataset.


\begin{table}[!tb]	
\centering
\addtolength{\tabcolsep}{0.5pt}
\renewcommand\arraystretch{1}
\setlength{\tabcolsep}{2.5pt}
\small
\begin{tabular}{ccc|ccc|cc}
\hline
\multicolumn{3}{c|}{Modules} & \multicolumn{3}{c|}{MOS}  & \multicolumn{2}{c}{EVE} \\  
Ra-OA & Ra-SA & Ra-CA & \textbf{mIoU}$\uparrow$ & \textbf{F1}$\uparrow$ & \textbf{mAcc}$\uparrow$ & \textbf{MAE}$\downarrow$ & \textbf{MSE}$\downarrow$\\

\hline 
\xmark & \xmark & \xmark & 65.7 & 72.9& 77.7 & 0.251 & 0.101 \\
\xmark & \xmark & \cmark & 66.9 & 73.8 & 80.3 & 0.227 & 0.093 \\
\cmark & \xmark & \cmark & 69.0& 75.5 & 80.2 & 0.197 &  0.071  \\           
\xmark & \cmark & \cmark & 67.5& 74.1 & 80.1 & 0.199 &  0.075  \\           
\cmark & \cmark & \xmark & 66.6& 73.8 & 79.1 & 0.198 &  0.072  \\
\cmark & \cmark & \cmark & \textbf{70.2} & \textbf{76.5} & \textbf{81.9} & \textbf{0.182} & \textbf{0.065}\\
\hline
\end{tabular}
\caption{Ablation study on MOS and EVE with each module}
\label{tab:4_5_ablantion_SA_VE}
\normalsize
\end{table}

\subsection{Ablation Study}
\textbf{Study on Radar Transformer.} The first ablation study validates the effectiveness of our proposed radar cross-attention (CA) and self-attention including object-attention (OA) and scenario-attention (SA) modules. We evaluate our method under different setups, including the one without all attention modules, while utilizing the original point self-attention module, the one only with CA, the one without CA but both self-attention modules, the one with one kind self-attention module and CA, as well as ours, using all attention modules. All setups use our velocity compensation. Ablation results for both MOS and EVE are shown in~\cref{tab:4_5_ablantion_SA_VE}. As shown, the setup with all attention modules (Ours) performs the best, while the performance significantly degraded if any module was missing. The biggest decline in performance was observed when all modules were disabled, indicating the importance of our proposed modules.


\begin{table}[!tb]
\centering
\addtolength{\tabcolsep}{0.5pt}
\renewcommand\arraystretch{1}
\setlength{\tabcolsep}{2.2pt}

\small
\begin{tabular}{cc|ccc|cc}
\hline
\multicolumn{2}{c|}{Velocity} & \multicolumn{3}{c|}{MOS}  & \multicolumn{2}{c}{EVE} \\  
uncalib-EVE & calib-EVE & \textbf{mIoU}$\uparrow$ & 
\textbf{F1}$\uparrow$ & \textbf{mAcc}$\uparrow$ & \textbf{MAE}$\downarrow$ & \textbf{MSE}$\downarrow$ \\
\hline
\xmark & \xmark & 61.1 & 69.3& 76.6 & 0.301 & 0.161 \\
\cmark & \xmark & 65.6 & 73.3 & 78.8 & \textbf{0.182} & \textbf{0.065} \\
\xmark & \cmark & \textbf{70.2} & \textbf{76.5} & \textbf{81.9} & - &  -  \\
\hline
\end{tabular}
\caption{Ablation study on MOS and EVE with velocity}
\label{tab:ablation_ve}
\end{table}

\noindent\textbf{Study on Velocity Compensation.} 
To validate the benefits of using velocity measurements and compensation for MOS and EVE, we conducted an ablation study to compare the performance with and without uncompensated radial velocity and velocity compensation. The results are presented in~\cref{tab:ablation_ve}. It shows that the performance gap between MOS models with and without using raw velocity measurements is large, with a 4.5\% improvement in mIoU. Compensation for radial velocity results in an additional 4.6\% improvement, leading to a total significant improvement of 9.1\% in mIoU. Furthermore, using raw velocity data from radar points improves the EVE results, reducing errors by approximately half for both MAE and MSE. Overall, our results demonstrate that our design of using radial velocity measurements and compensation significantly improves the performance of our method for both MOS and EVE.

Additionally, We investigated the impact of the Doppler loss for EVE, the influence of the ball radius in OA, and the effects of the k value in the kNN of SA for MOS, respectively. We also explored different training strategies of multi-task multi-head networks for radar MOSEVE tasks. Due to page limitations, the details of these ablations studies are presented in the supplementary materials.

\section{Conclusion}
In this paper, we introduced a novel radar transformer network to address both MOS and EVE tasks using sparse radar point clouds. Our approach is based on our proposed radar self- and cross-attention mechanisms, which can effectively extract distinctive features from sparse radar points. Additionally, our method estimates the radial velocity of radar point clouds and utilizes it to compensate for the raw radar velocity measurements. Finally, it takes two compensated radar point clouds as input to generate the MOS results.
We evaluate the performance of different methods on radar MOS and EVE tasks in our dataset and the VoD dataset. The experimental results demonstrate that our proposed method outperforms existing state-of-the-art methods in both tasks, obtaining improvements of more than 17\,\% in EVE precision and 9.4\,\% in MOS mIoU. We release the implementations of our method and dataset with the annotations to facilitate future research on Radar MOSEVE tasks.

\noindent\textbf{Limitation.} Although our method achieves good performance, there are still complex scenarios where it struggles to perform well. For instance, when the scene contains only moving objects, and the sparse radar measurements are solely on these objects, it creates an ill-defined MOSEVE problem, particularly when the object's motion is similar to that of the robot. One potential solution is to incorporate additional sensor modalities, such as images and LiDAR, to provide more environmental measurements and improve MOSEVE performance in such challenging situations.

\noindent\textbf{Societal Impacts.} Our approach is capable of accurately estimating the ego velocity and detecting potentially moving objects, such as pedestrians, in driving environments. This is particularly important for safety-critical real-world applications, such as autonomous cars and mobile robots.

\section*{Acknowledgments}
This work was partly supported by the ORCA-UBOAT company and the National Science Foundation of China under Grant U1913202, U22A2059, and 62203460, as well as the Natural Science Foundation of Hunan Province under Grant 2021JC0004 and 2021JJ10024.

\bibliography{aaai24}

\end{document}